\documentclass{article}

\usepackage[preprint]{neurips_2019}

\usepackage[utf8]{inputenc} 
\usepackage[T1]{fontenc}    
\usepackage{hyperref}       
\usepackage{url}            
\usepackage{booktabs}       
\usepackage{amsfonts}       
\usepackage{nicefrac}       
\usepackage{microtype}      
\usepackage{graphicx}
\usepackage{algorithmicx}
\usepackage{algorithm}
\usepackage[noend]{algpseudocode}
\usepackage{amsmath}
\usepackage{natbib}
\title{HUGE$^2$: a \underline{H}ighly \underline{U}ntangled \underline{G}enerative-model \underline{E}ngine for \underline{E}dge-computing}

\author{
  Feng Shi, Ziheng Xu, Tao Yuan, Song-Chun Zhu \\
  Department of Computer Science\\
  University of California\\
  Engineering VI, Los Angeles, CA 90095 \\
  \texttt{shi.feng@cs.ucla.edu, lawrencexu@g.ucla.edu} \\
}

\begin{document}

\maketitle

\begin{abstract}
  As a type of prominent studies in deep learning, generative models have been widely investigated in research recently. Two research branches of the deep learning models, the \textit{Generative Networks} (GANs, VAE) and the \textit{Semantic Segmentation}, rely highly on the upsampling operations, especially the \textit{transposed convolution} and the \textit{dilated convolution}.
  However, these two types of convolutions are intrinsically different from standard convolution regarding the insertion of zeros in input feature maps or in kernels respectively. This distinct nature severely degrades the performance of the existing deep learning engine or frameworks, such as Darknet, Tensorflow, and PyTorch, which are mainly developed for the standard convolution. Another trend in deep learning realm is to deploy the model onto edge/ embedded devices, in which the memory resource is scarce. In this work, we propose a \underline{H}ighly \underline{U}ntangled \underline{G}enerative-model \underline{E}ngine for \underline{E}dge-computing or \textit{HUGE$^2$} for accelerating these two special convolutions on the edge-computing platform by decomposing the kernels and untangling these smaller convolutions by performing basic matrix multiplications. The methods we propose use much smaller memory footprint, hence much fewer memory accesses, and the data access patterns also dramatically increase the reusability of the data already fetched in caches, hence increasing the localities of caches. Our engine achieves a speedup of nearly 5$\times$ on embedded GPUs, and around 10$\times$ on embedded CPUs, and more than 50\% reduction of memory access.
\end{abstract}


\section{Introduction}
Recently, the deep generative models and semantic segmentation algorithms have shown their stunning abilities in various fields, such as creating realistic images from the learned distribution of a given dataset, providing the robots with the ability to learn from environment without human input, generating the synthetic 3D objects for the scene parsing in a scenario, and so forth. These creative deep learning models attract great interests in research by both scholar and industry. The representative works include the Generative Neural Networks (GAN) \cite{Goodfellow}, the Variational Auto-encoder (VAE) \citet{KingmaW13}, and the semantic image segmentation algorithms \citet{7478072, ChenPSA17}.

However, the generative models and semantic segmentation algorithms rely heavily on the \textit{deconvolution} which is an inefficient, and both computation- and memory-intensive operation. The inefficiency comes either from the zero insertions in either input tensor or kernels or from repeatedly accesses to the overlapped regions. Zero insertions cause wasteful computations, hence high latency. The non-consecutive memory access manner in deconvolutions also hurts system performance drastically. The overlapped region in outputs hinders the concurrent processing because the chained memory-writings happen to the same location.

In this work, we conceive a set of solutions from an algorithmic perspective to improve the performance of deconvolution for embedded systems. 
And the experiments show that we achieve the speedup nearly 10$\times$ on CPUs and 5$\times$ on GPU, the memory storage and their accesses are reduced by more than 50 percent.



\section{Background}\label{sec:background}
\begin{figure*}[ht]
  \centering
  \includegraphics[width=0.60\textwidth]{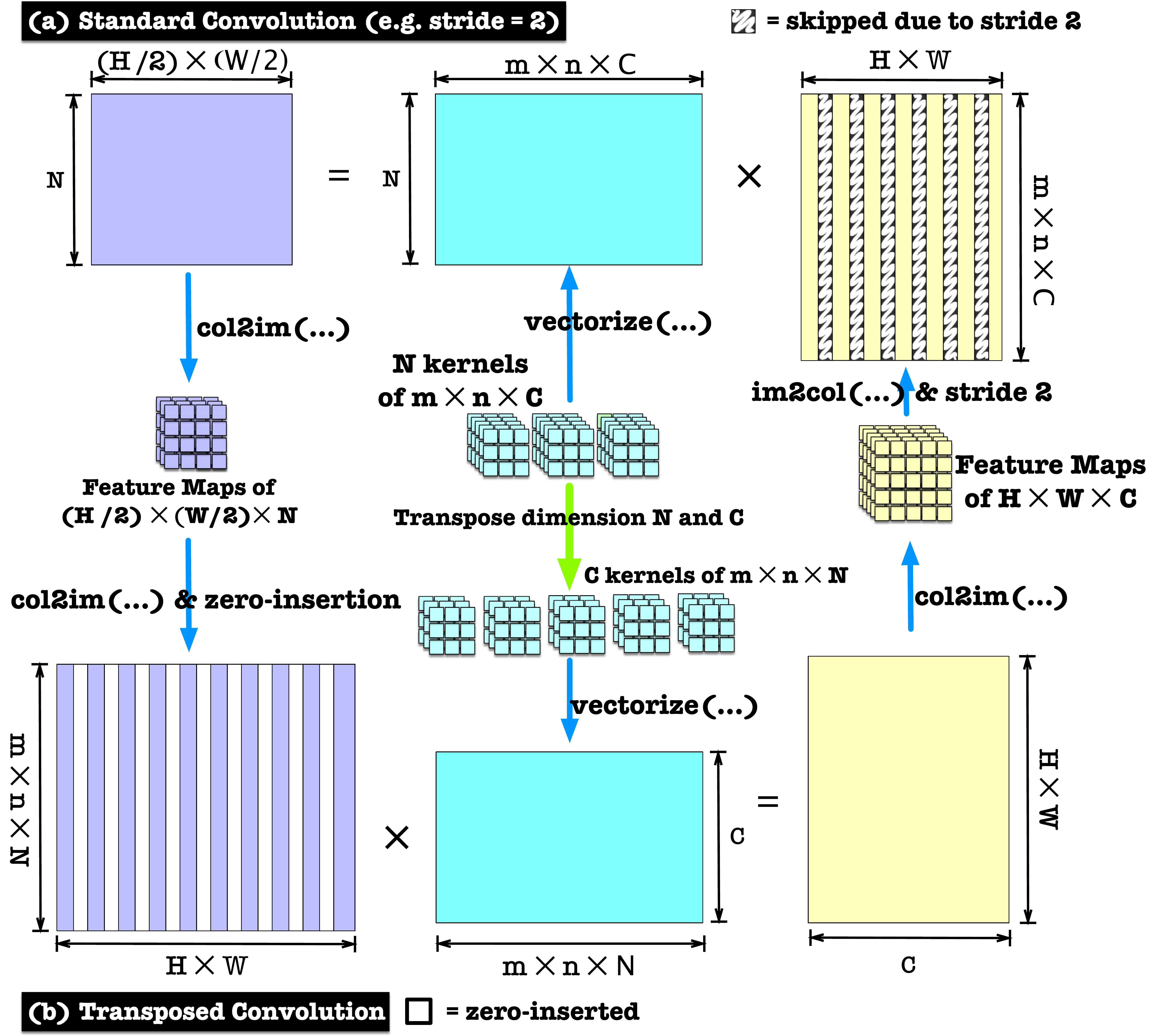}
  \caption{The upper is the implementation of a strided convolution, and the bottom is its related transposed convolution.}
  \label{fig:contrerpart}
\end{figure*}
\subsection{Deconvolution}
The so-called \textbf{deconvolution} used in the deconvolution layers of the \textit{generative adversarial networks} and the \textit{semantic image segmentation} is actually not as exact as the reverse operation of the \textit{convolution}.
Actually, the deconvolution layers are learnable up-sampling layers. Two categories of special convolution operations can fulfill such kind of task, they are the \textit{transposed convolution} and the \textit{dilated convolution}, respectively.
The following subsections give details about how these operations work.

\subsubsection{Transposed Convolution}


\textit{Transposed convolution}, also called \textit{Fractionally-Strided Convolution}, is used not only to upsample an initial layer but also to create new features in enlarged output feature maps. Theoretically, transposed convolution works as a process of swapping the Forward and backward passes of a convolution, and this is where its name comes from. Algorithm \ref{alg:t_conv_fv} describes how this kind of convolution works. As it shows,  when $s_{m}$ and $s_{n}$ are bigger than 1, the kernels slide on the feature maps with fractional steps. Figure \ref{fig:contrerpart} shows the implementations of the transposed convolution and its counterpart.



\begin{algorithm}[h]
  \centering
  \caption{Transposed Convolution}
  \label{alg:t_conv_fv}
  \begin{algorithmic}
    \Function{Conv2DTranspose}{$I, K, O, s_m, s_n$}
    \Comment{$s_m$, $s_n$ are fraction factor also zero-insertion stride on input tensors}
      \For{$0 \leq k \leq N - 1$}
        \For{$0 \leq h \leq H - R + 1$}
          \For{$0 \leq w \leq W- S + 1$}
            \For{$0 \leq c \leq C - 1$}
              \For{$0 \leq m \leq R - 1, m = m + s_{m} $}
                \For{$0 \leq n \leq S - 1, n = n + s_{n} $}
                  \State $O[h, w, k] $ += $I[h + \frac{m}{s_{m}}, w + \frac{n}{s_{n}}, c] \times K[h \bmod s_{m} + m, w \bmod s_{n} + k_{n}, c, k]$
                \EndFor
              \EndFor
            \EndFor
          \EndFor
        \EndFor
      \EndFor
    \EndFunction
  \end{algorithmic}
\end{algorithm}


\begin{algorithm}[h]
  \centering
  \caption{Dilated Convolution}
  \label{alg:d_conv_v}
  \begin{algorithmic}
    \Function{Conv2DDilated}{$I, K, O, s_m, s_n$} \Comment{$s_m$, $s_n$ are dilation factors also zero-insertion stride on kernels}
      \For{$0 \leq k \leq N - 1$}
        \For{$0 \leq h \leq H - R + 1$}
          \For{$0 \leq w \leq W - S + 1$}
            \For{$0 \leq c \leq C - 1$}
              \For{$0 \leq m \leq R - 1$}
                \For{$0 \leq n \leq S - 1$}
                  \State $O[h, w, k]$ += $I[h + s_{m} \times m, w + s_{n} \times n, c] \times K[m, n, c, k]$
                \EndFor
              \EndFor
            \EndFor
          \EndFor
        \EndFor
      \EndFor
    \EndFunction
  \end{algorithmic}
\end{algorithm}

\begin{figure}[ht]
  \centering
  \includegraphics[width=0.48\textwidth]{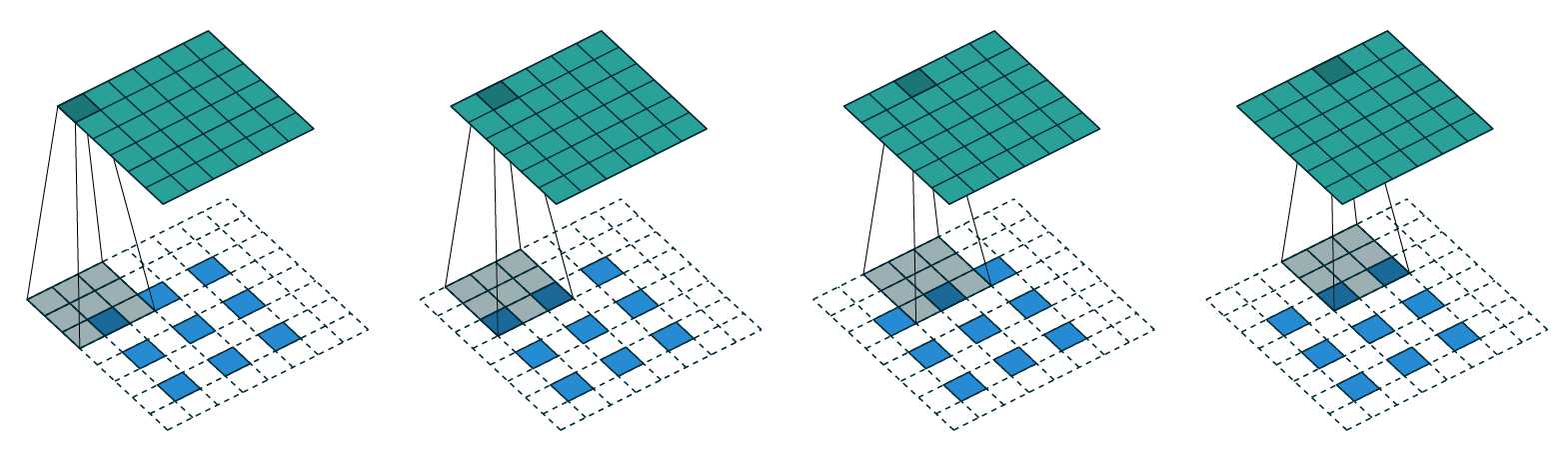}
  \includegraphics[width=0.48\textwidth]{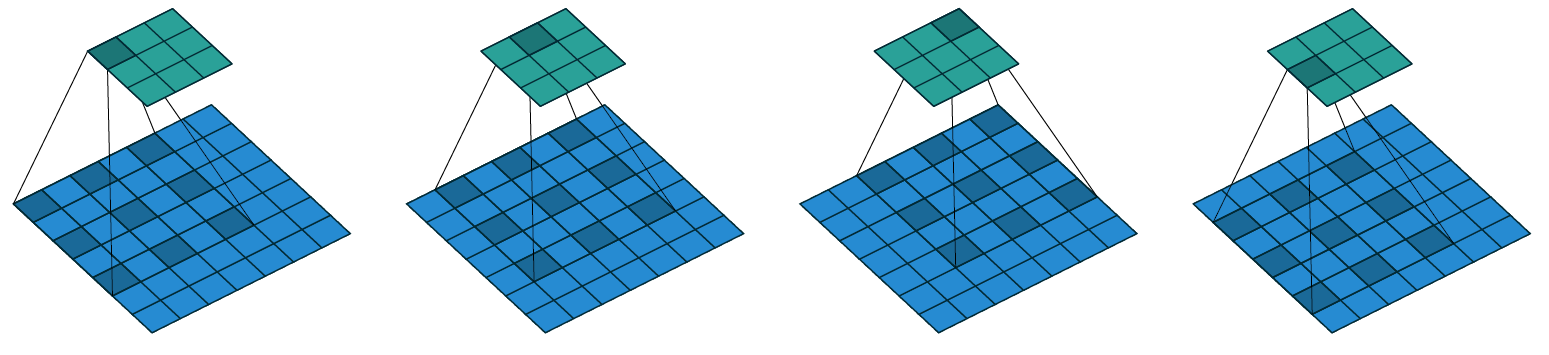}
  \caption{left: the transposed convolution of a $3 \times 3$ kernel over a $6 \times 6$ input padded with a $1 \times 1$ border of zeros using $2 \times 2$ strides; right: dilated convolution of a $3 \times 3$ kernel over a $7 \times 7$ input with a kernel of the dilation factor of 2 \cite{2016arXiv160307285D}.}
  \label{fig:transpose_scene}
\end{figure}

It is always possible to emulate a transposed convolution with a direct convolution. Such process first spreads the input feature map by inserting zeros (or blank lines) between each pair of rows and columns. The original input tensor $I$ now becomes $\hat{I}$. It then applies a standard convolution, with strides of 1 (equivalent to sliding step of $\frac{1}{stride}$ on the original input \cite{2016arXiv160307285D}), on the resulting input representation, as shown in Algorithm \ref{alg:t_conv_fv}. Let us take left hand side of Figure \ref{fig:transpose_scene} as an example. It illustrates a $2D$ transposed convolution of a zero-inserted feature map of size $6 \times 6$ and a $3 \times 3$ transposed kernel.
\subsubsection{Dilated Convolution}

%
%


Strictly speaking, despite of that dilated convolution, also known as \textit{atrous convolution}, is not acknowledged as a kind of deconvolution, it has been widely explored to upsample input tensors in the semantic image segmentation algorithms. Moreover, it shares some characteristics on which we can apply our acceleration algorithm as it does for the transposed convolution. On the contrary to transposed convolution, dilated convolution inserts zeros into kernels but not input tensors. The kernels are dilated so as to enlarge their corresponding receptive fields. One thing needs to be noticed is that only stride bigger than 1 has the effect of upsampling on input tensors. Details are demonstrated in Algorithm \ref{alg:d_conv_v} and right side of Figure \ref{fig:transpose_scene}. 

\subsection{Previous Work}
To our best knowledge, up to this work, most of optimized solutions have been proposed are from the research of the hardware accelerator realm. These designated designs achieve much higher throughput compared with non-optimized generic hardware. Hence, our goal is to conceive an easily-accessible and cost-efficient solution for the generic hardware.
\begin{enumerate}
    \item \textbf{Zero-Skipping}: \cite{GANAX, FlexiGAN} present a set of designs by swapping zero rows and columns with non-zeros ones, and then rearranging non-zero rows and columns into effective working groups. However, this design doesn't thoroughly solve the unbalanced working load problem among effective computation groups. \cite{8326999} discovers the delicate mathematical relation of indices among input tensors, kernels, and output tensors. These relations help rearrange the computations to skip zeros. But this method lacks memory access coalescing. Therefore, input tensors and kernels are accessed in a non-consecutive fashion with degradation in the overall performance of the system.
    \item \textbf{Reverse Looping and Overlapping}: \textit{Reverse Looping} is introduced by both \cite{Zhang2017ADM} and \cite{Xu_fcn}. This technique avoids accessing the output tensors in an overlapped manner with more operations, especially the accumulations and memory writings. Reverse looping, on the contrary, uses the output space to determine corresponding input blocks, and thus eliminating the need for the additional accumulations and memory accesses. However, the overlapped regions are not evenly distributed, hence the work load unbalancing issue among processing elements is still not well solved by such kind of solution.
\end{enumerate}
\section{Algorithm} \label{sec:algorims}
In this section, we introduce our algorithm for accelerating the deconvolution operations. Our algorithm consists of three steps: 1) kernel decomposition, 2) untangling of kernels and matrix multiplications, and 3) dispatching and combining the results to the output tensor.

The following subsections provide the explanation for each of them.

\begin{figure}[ht]
    \centering
    \includegraphics[width=0.6\textwidth]{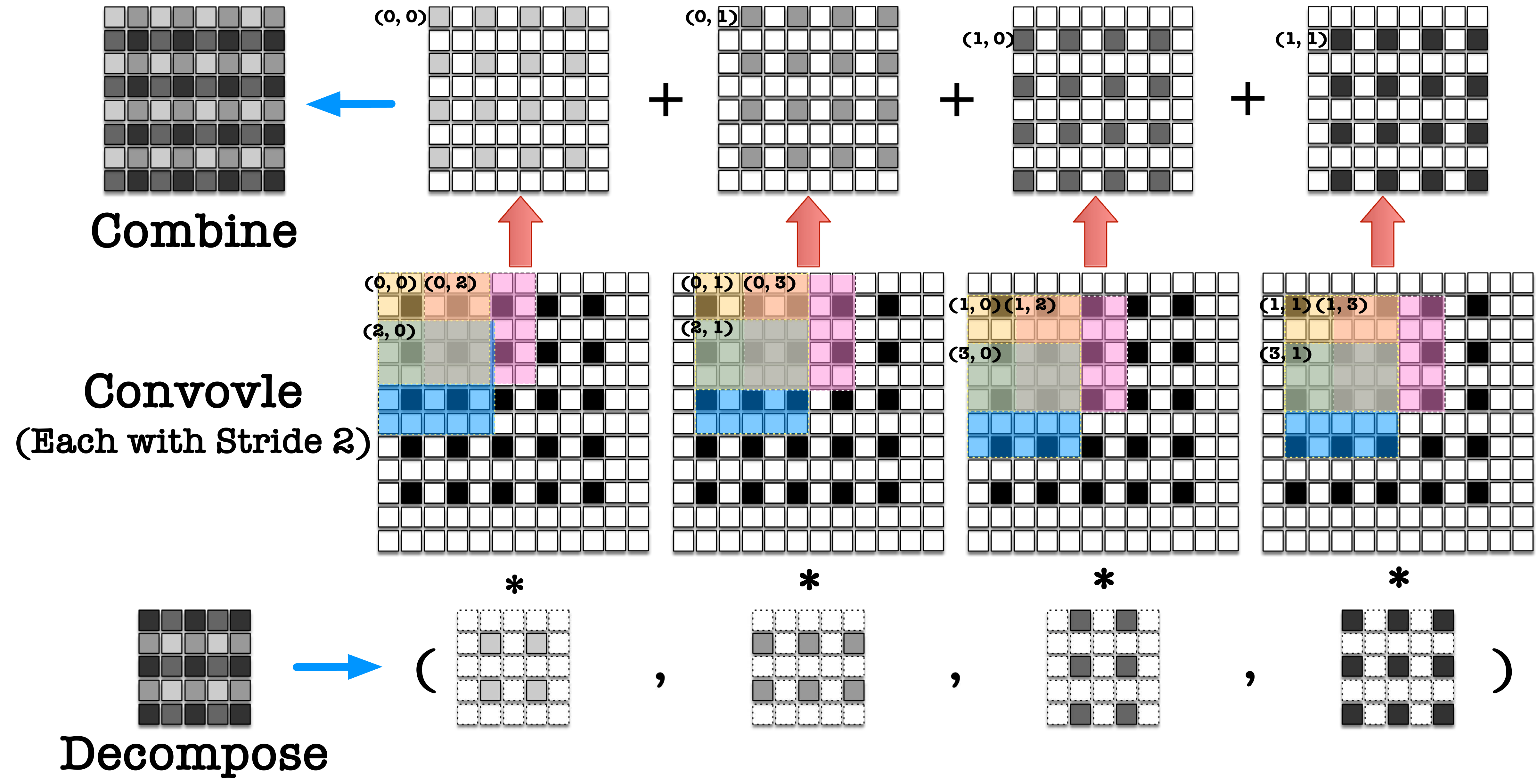}
    \caption{the kernel of the transpose convolution is decomposed into 4 patterns, each convolves with zero-inserted feature maps with stride 2, the final result is obtained by combining the 4 partial feature maps; the yellow, pink, and blue patches correspond to sliding windows at different positions.}
    \label{fig:patterns}
\end{figure}


\subsection{Decompose Deconvolution}
Given an input tensor with stride 2 zero-inserted as example, and let us take a transposed kernel to slide on it. We discover that there exists 4 kinds of patterns as shown at the bottom of Figure \ref{fig:patterns} where the nonzero elements in the kernel meet the nonzero elements in the zero-inserted input tensor $\hat{I}$, and thus generate non-overlapped effective outputs as shown on the top of Figure \ref{fig:patterns}. Mathematical description of these 4 patterns is given below:

  \textbf{Pattern 1}: \textit{odd columns} and \textit{odd rows} of kernel convolve with stride 2 on input tensor $\hat{I}$ and generate \textit{even columns} and \textit{even rows} of output tensor $O$.
  \begin{equation}
    O[2h, 2w, k] = \sum_{c=0}^{C} \sum_{m=0}^{R} \sum_{n=0}^{S} K[2m + 1, 2n + 1, k, c] \times \hat{I}[2h + 2m, 2w + 2n, c]
  \end{equation}

  \textbf{Pattern 2}: \textit{even columns} and \textit{odd rows} of kernel convolve with input tensor $\hat{I}$ and generate \textit{odd columns} and \textit{even rows} of output tensor $O$.
  \begin{equation}
    O[2h, 2w + 1, k] = \sum_{c=0}^{C} \sum_{m=0}^{R} \sum_{n=0}^{S} K[2m + 1, 2n, k, c] \times \hat{I}[2h + 2m, 2w + 2n + 1, c]
  \end{equation}

  \textbf{Pattern 3}: \textit{odd columns} and \textit{even rows} of kernel convolve with input tensor $\hat{I}$ and generate \textit{even columns} and \textit{odd rows} of output tensor.
  \begin{equation}
    O[2h + 1, 2w, k] = \sum_{c=0}^{C} \sum_{m=0}^{R} \sum_{n=0}^{S} K[2m, 2n + 1, k, c] \times \hat{I}[2h + 2m + 1, 2w + 2n, c]
  \end{equation}

  \textbf{Pattern 4}: \textit{even columns} and \textit{even rows} of kernel convolve with input tensor $\hat{I}$ and generate \textit{odd columns} and \textit{odd rows} of output tensor.
  \begin{equation}
    O[2h + 1, 2w + 1, k] = \sum_{c=0}^{C} \sum_{m=0}^{R} \sum_{n=0}^{S} K[2m, 2n, k, c] \times \hat{I}[2h + 2m + 1, 2w + 2n + 1, c]
  \end{equation}


\begin{figure}[ht]
    \centering
    \includegraphics[width=0.60\textwidth]{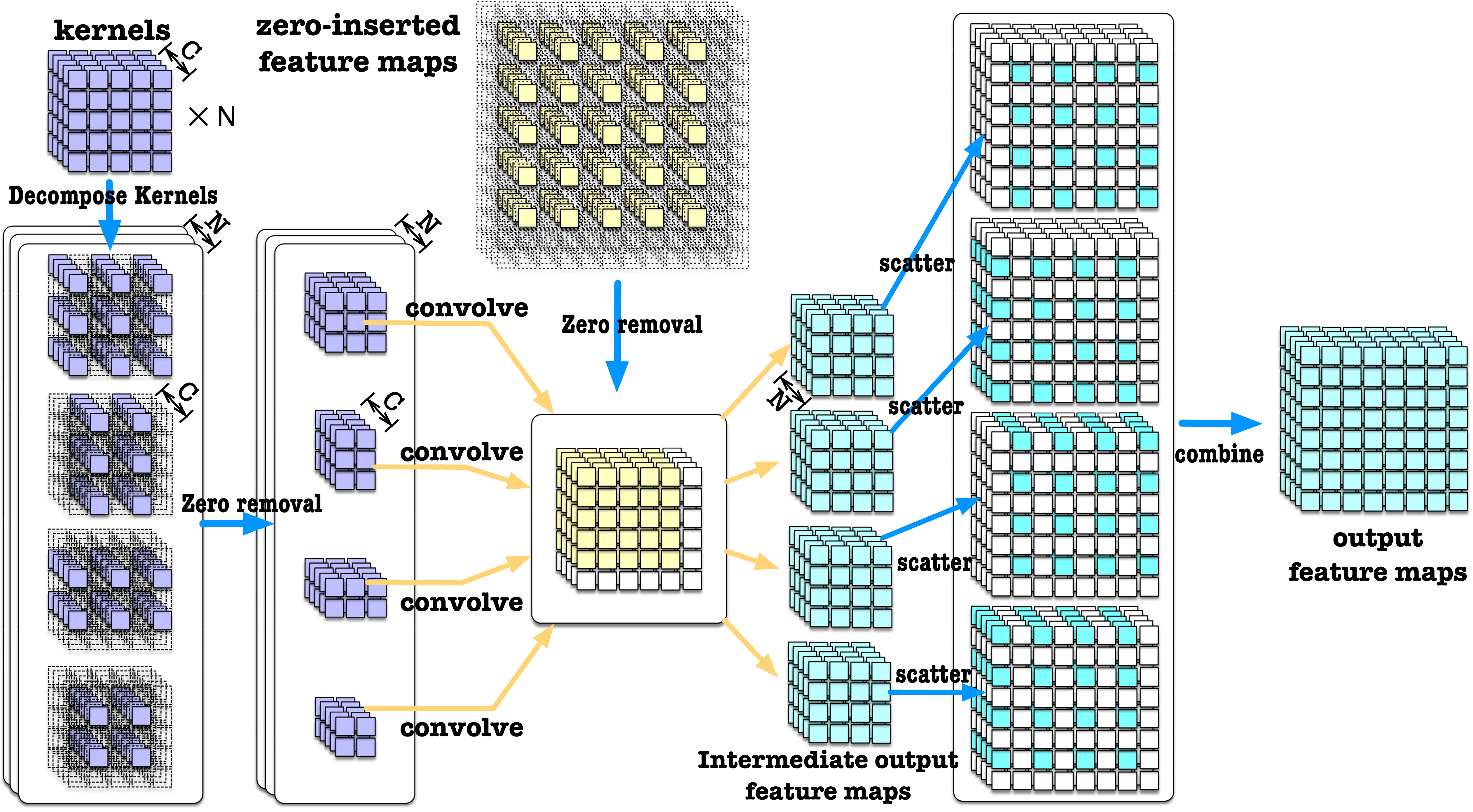}
    \includegraphics[width=0.36\textwidth]{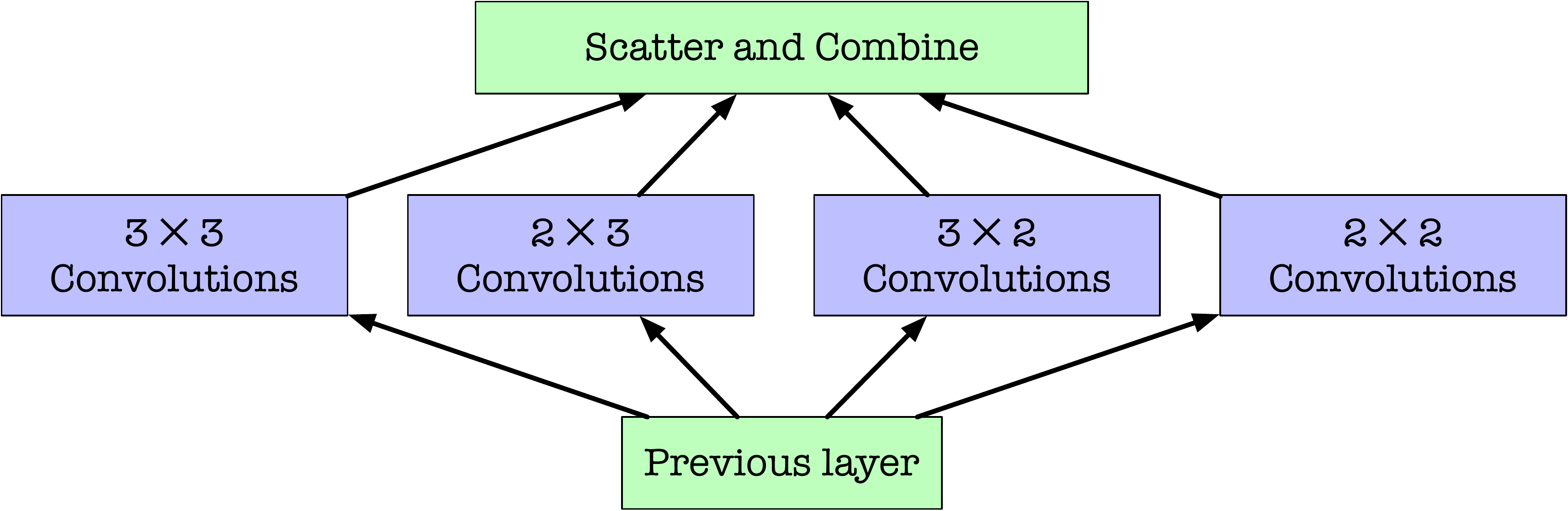}
    \caption{left: the corresponding flow of the example; right: a simplified view}
    \label{fig:flow}
\end{figure}

One benefit of such decomposition is that the non-overlapped sparse regions on the output tensor do not cause any race conditions for writing in memory, since it will not be blocked by consecutive memory writings.

After having investigated the relation between indices of the nonzeros in input tensor and the decomposed kernels, we draw the conclusion that we can safely remove all the zero inserted in both input tensor and the decomposed kernels as shown in left side of Figure \ref{fig:flow}. The flow demonstrates the zero-removal for all patterns where they become 4 smaller standard convolutions on input tensors without zero-insertion. Then we scatter and combine their results. The scattering of the results from each pattern follows the corresponding indices used in the zero-inserted version.

And the simplified flow on the right side of Figure \ref{fig:flow} resembles the Inception module of GoogLeNet \cite{43022} except that the last step here is scattering and combination instead of stacking.






\subsection{Untangling}


\begin{figure*}[ht!]
  \centering
  \includegraphics[width=0.7\textwidth]{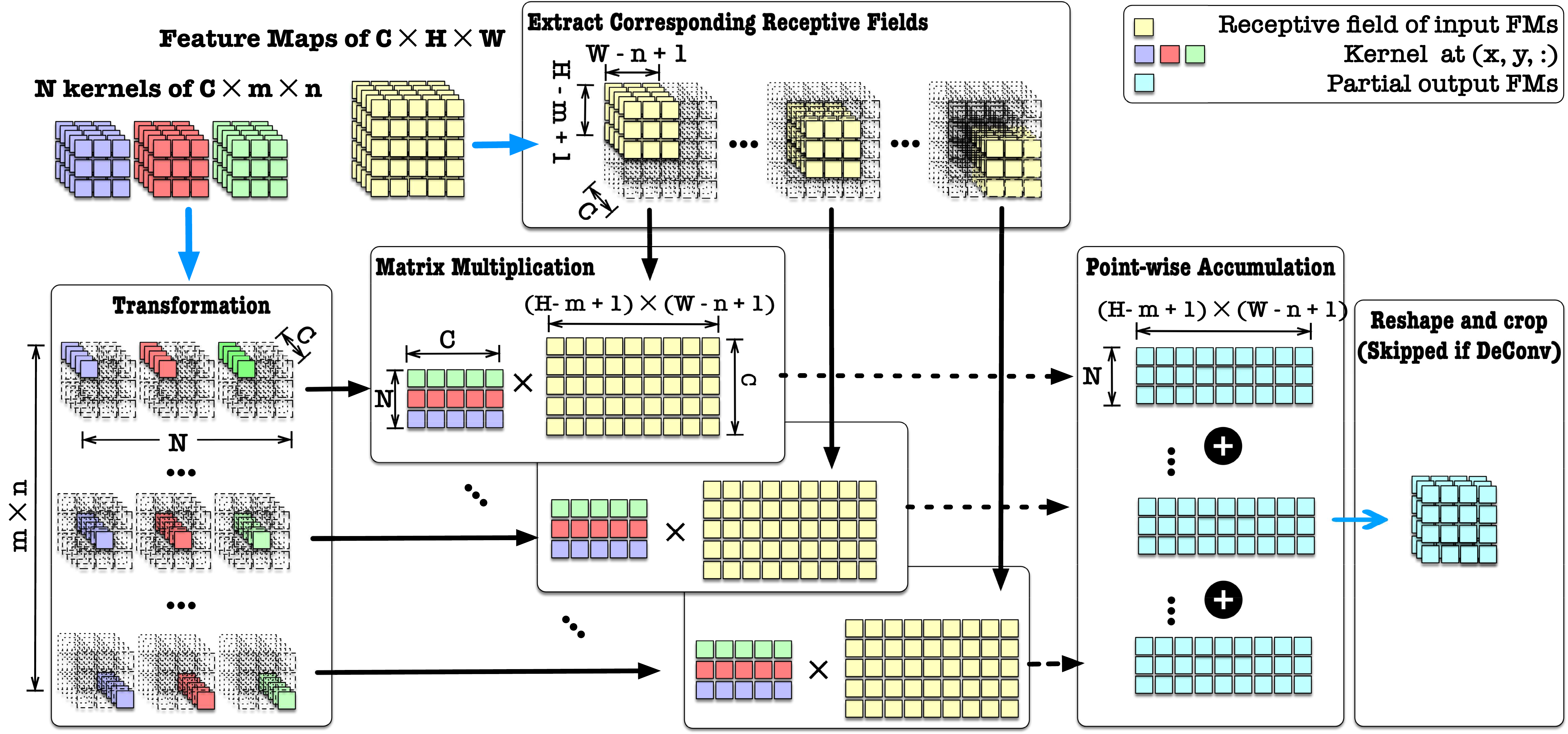}
  \caption{untangle a standard convolution into a set of 1$\times$1 convolutions}
  \label{fig:untangle_conv2d}
\end{figure*}
To further improve the parallelism of arithmetic computations, we propose an algorithm to untangle every decomposed pattern of the transposed convolution into a set of $1 \times 1$ convolutions.

\subsubsection{Untangle standard convolution}

As shown in the right side of Figure \ref{fig:flow} each pattern convolves with input tensor as a standard convolution. To better understand our algorithm, let us take pattern 4 (the $ 3\times 3$ one) as an example. The process is shown in Figure \ref{fig:untangle_conv2d}.
Given $N$ decomposed kernels of pattern 4 (with zero removed) from previous subsection and the original input tensor, each kernel has dimension of $m \times n \times C$ and the input tensor has dimension $H \times W \times C$. We regroup the elements of the kernels by gathering $N$ columns along the dimension $C$ from every kernel at position $(x, y)$ (e.g. $(0, 0)$ at top left case, $(0, 2)$ for the bottom right case in the Figure \ref{fig:untangle_conv2d}) of the $(m, n)$ plane. These columns form a matrix of dimension $N \times C$. Then, their corresponding receptive fields on the input tensor can be fetched for input tensor to form another matrix of dimension $(H - m + 1)(W - n + 1) \times C$. Such configuration can be regarded as a $1 \times 1$ convolution with $N$ $1 \times 1$ kernels working with a cropped tensor.
The products of the $m \times n$ matrix multiplications are then accumulated together. The elements of the resultant matrix are then dispatched to the corresponding position in the output tensor.

\subsubsection{Untangle dilated convolution}

\begin{figure*}[ht!]
  \centering
  \includegraphics[width=0.48\textwidth]{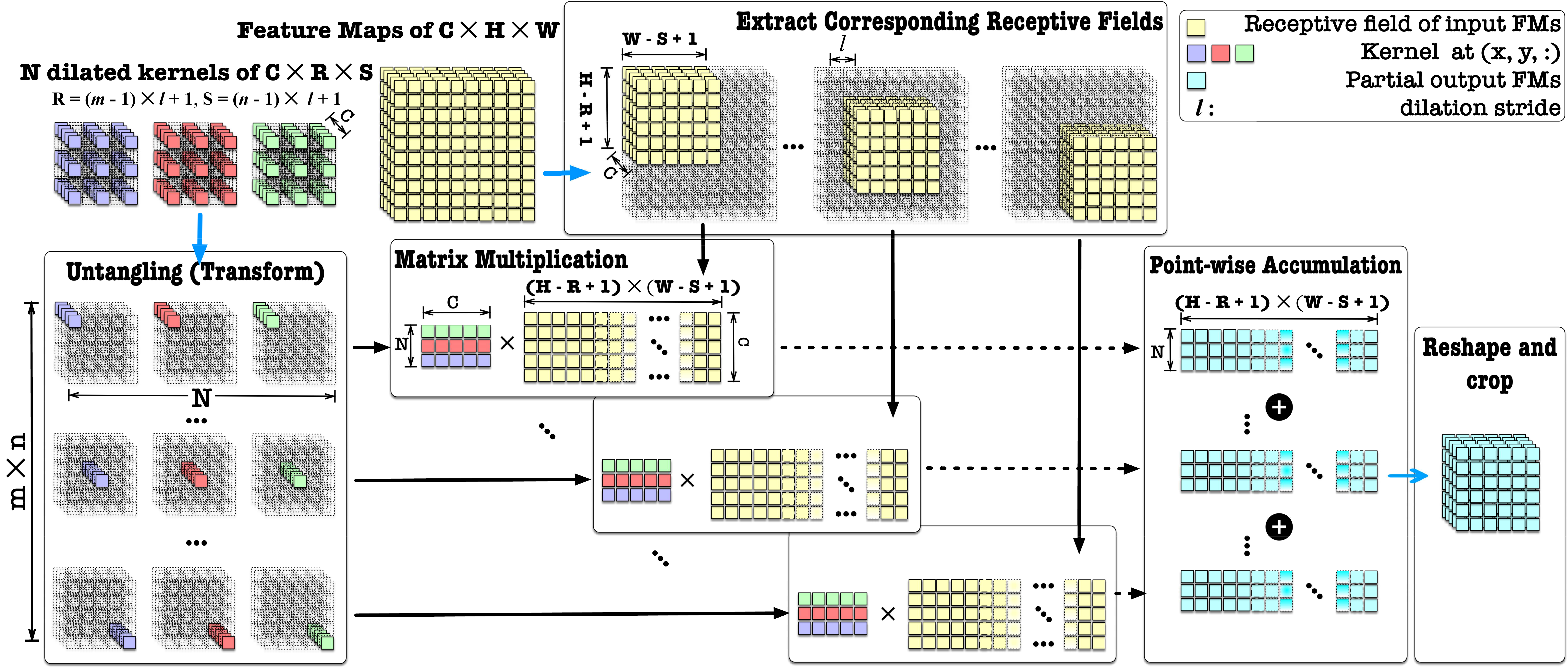}
  \includegraphics[width=0.48\textwidth]{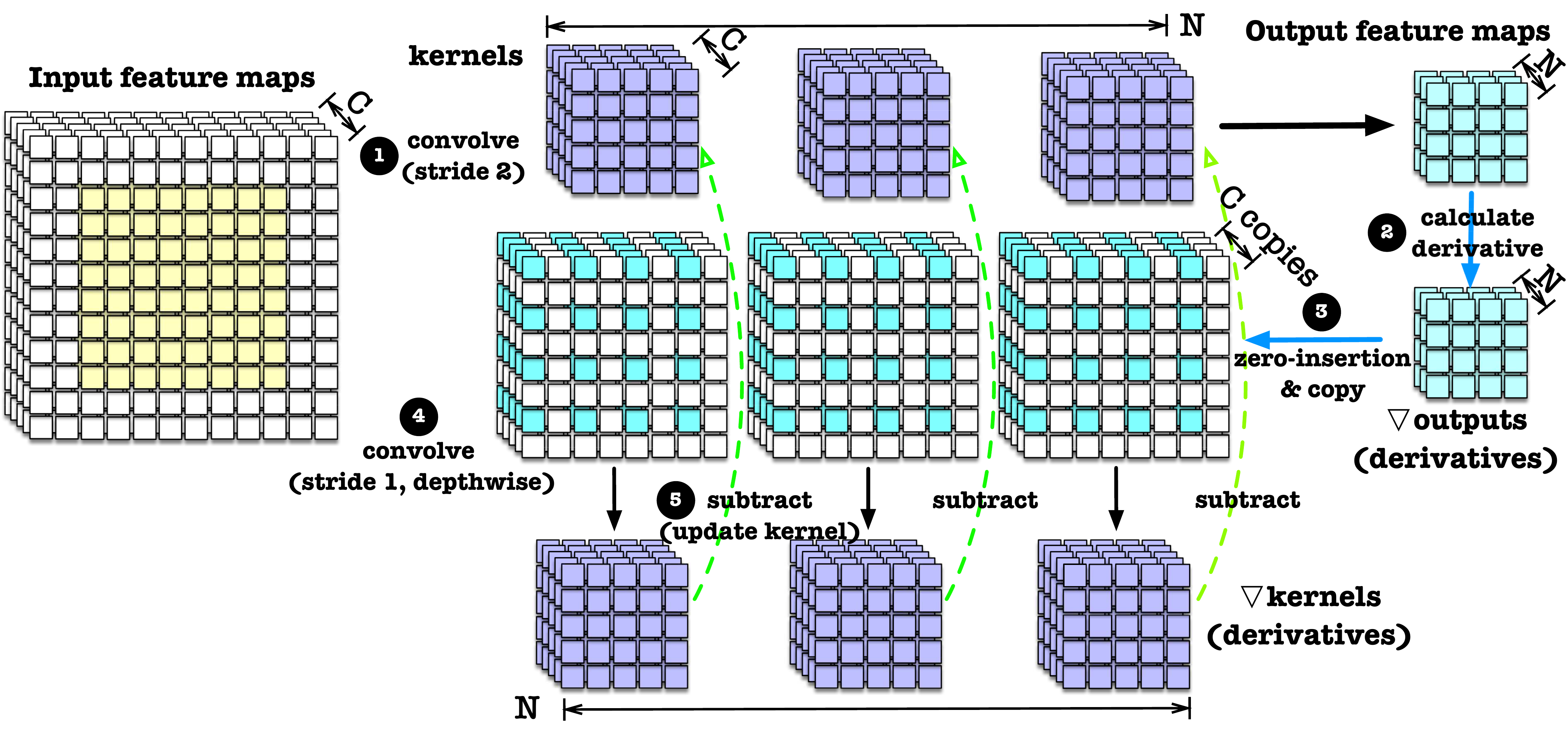}
  \caption{left: untangle a dilated convolution; right: training of discriminator with dilated convolutions}
  \label{fig:untangle_dilate}
\end{figure*}
Dilated convolution can also take the advantage of untangling. As it shows in left side of Figure \ref{fig:untangle_dilate} untangling technique is also applicable. The sliding step on input tensor is larger, and the receptive field shrink with multiple of stride.

\subsubsection{Training of GAN}
The back-propagation of the discriminator of GAN can be seen as a special case of dilated convolution. Step 3 of right side of Figure \ref{fig:untangle_dilate} depicts that, in order to propagate the derivatives of the output errors, there are $C \times N$ convolutions by $C$ input feature maps and $N$ derivative maps. Each derivative map is dilated (since discriminator uses the strided convolution) and convolves with each input feature map.

Therefore, we can make $C$ copies of each of N derivative maps from output errors to form $N$ new dilated kernels of $C$ channels. Then the dilated kernels convolve with input tensor to form the derivates of kernels and the results are subtracted from corresponding kernels.

The dilated convolution in step 4 of right side of Figure \ref{fig:untangle_dilate} is actually a depth-wise version. Hence, it corresponds to $C=1$ in left side of Figure \ref{fig:untangle_dilate} which is seen as a outer-product of two vectors.

The back-propagation of generator of GAN can be seen as a strided convolution of derivative maps of output errors and input tensor (not shown in this paper).

\section{Experimental Results} \label{sec:experiments}
This section provides the evaluation of our algorithms. We use the deconvolution layers of DCGAN \cite{RadfordMC15} and cGAN \cite{MirzaO14} as case study. Their configurations are shown in Table \ref{tab:config}. In this paper, we mainly focus on the inference phase of deconvolution layers, and all models are pretrained with CIFAR100 \cite{cifar} dataset. The experiments for GAN's training are only investigated at several typical layers.
\begin{table}[h]
  \centering
  \caption{Configuration of deconvolution layers}
  \small{
  \begin{tabular}[t]{llcccc}
    \toprule
    \textbf{GAN} & \textbf{Layer} & \textbf{Input} & \textbf{Kernel} & \textbf{Stride}\\
    \midrule
    DCGAN & DC1 & $4\times4\times1024$  & $5\times5\times1024, 512$ & $2\times2$ \\
          & DC2 & $8\times8\times512$   & $5\times5\times512, 256$ & $2\times2$  \\
          & DC3 & $16\times16\times256$ & $5\times5\times256, 128$ & $2\times2$  \\
          & DC4 & $32\times32\times128$ & $5\times5\times128, 3$   & $2\times2$  \\
    CGAN  & DC1 & $8\times8\times256$  & $4\times4\times256, 128$ & $2\times2$ \\
          & DC2 & $16\times16\times128$   & $4\times4\times128, 3$ & $2\times2$  \\
    \bottomrule

  \end{tabular}
  }
  \label{tab:config}
\end{table}

The baseline of library we pick up is DarkNet \cite{darknet13} since it is open-sourced, and the commercial library such as cuDNN \cite{cudnn} only delivers with binary code.
The system used in our experiments equips with embedded CPU, 4-core ARM Cortex-A57, and a Nvidia's GPU (256-core NVIDIA Pascal™ Embedded GPU). The experiments are run on both embedded CPU and embedded GPU. The metrics used for performance comparison includes the speedup and memory access reduction. We compared our implementations of transposed convolution and dilated convolution with the baseline, which is the naive implementations from DarkNet for both CPU and GPU. Most 2D standard and transpose convolution implementation in modern deep learning library are based on \textbf{im2col}.

\subsection{Speedup of Computation}
\begin{figure}[ht]
  \centering
  \includegraphics[width=0.48\textwidth]{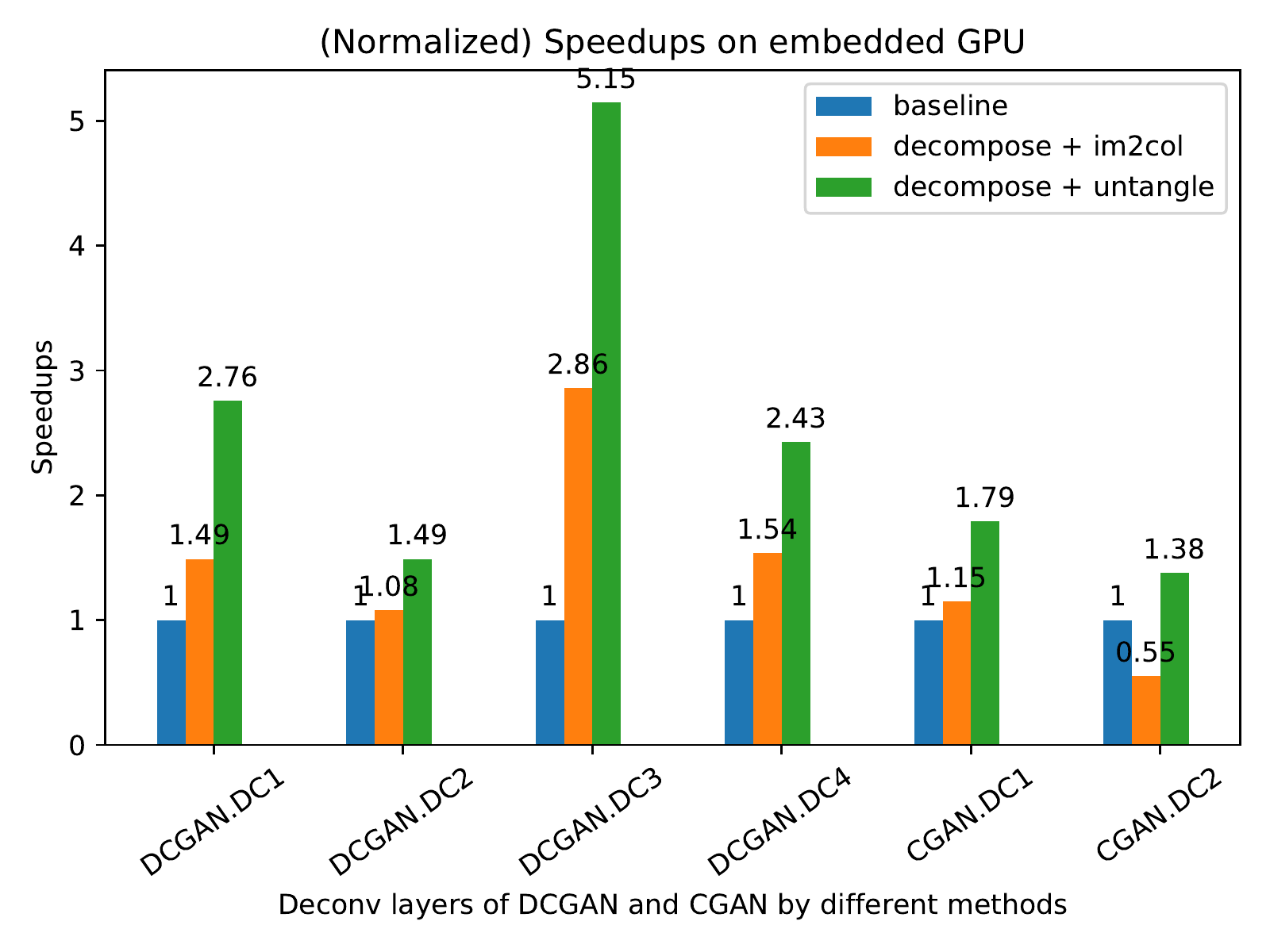}
  \includegraphics[width=0.48\textwidth]{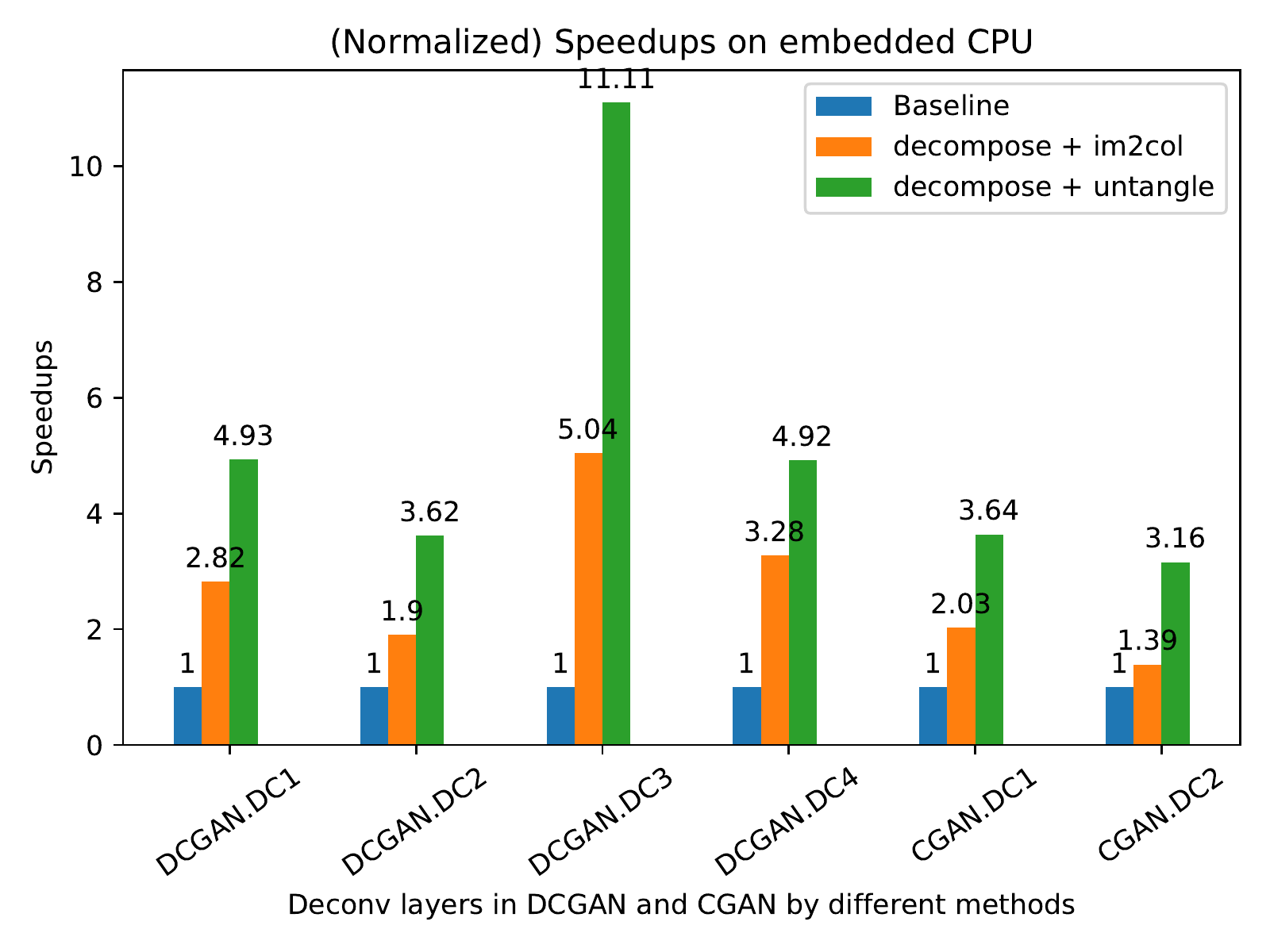}
  \caption{The speedups of the inference of GANs on embedded system. Left: on embedded GPU of Jetson TX2; right: on CPU (4-core Cortex-A57) of the same board}
  \label{fig:sp_gan}
\end{figure}

The speedup is obtained by the comparison of the computational runtime with the baseline. Figure \ref{fig:sp_gan} demonstrates the speedup gained by applying kernel decomposition and untangling for DCGAN and cGAN, respectively.

From the results, we can see that the shallower deconvolution layer are more compute-bounded since they have more kernels which deconvolve with input tensor and require much more computational operations. Untangling transposed kernels can efficiently improve the parallelism by taking advantage of larger $C$ and $N$.

\subsection{Reduction of Memory Access}
\begin{figure}[ht]
  \centering
  \includegraphics[width=0.48\textwidth]{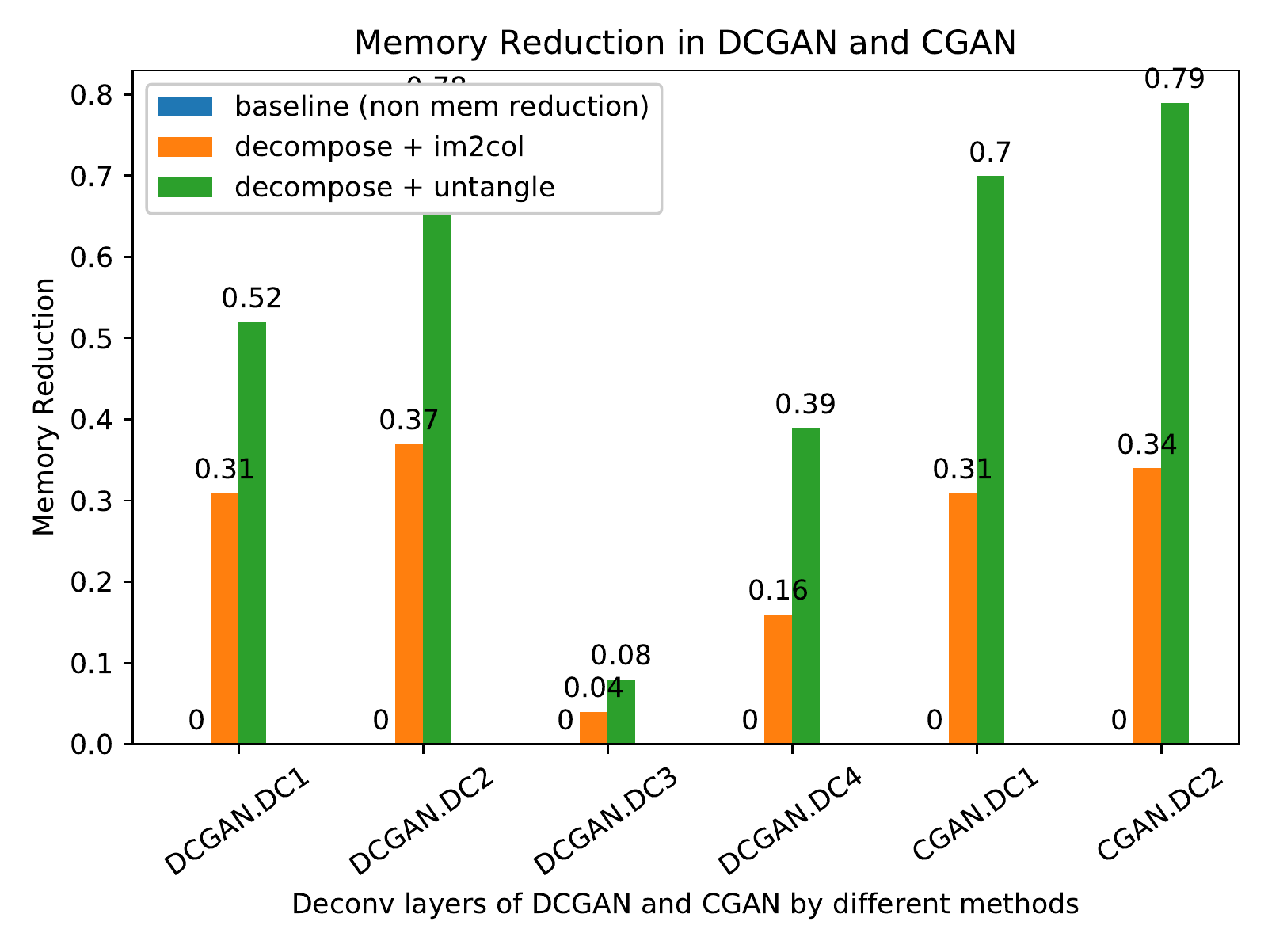}
  \includegraphics[width=0.48\textwidth]{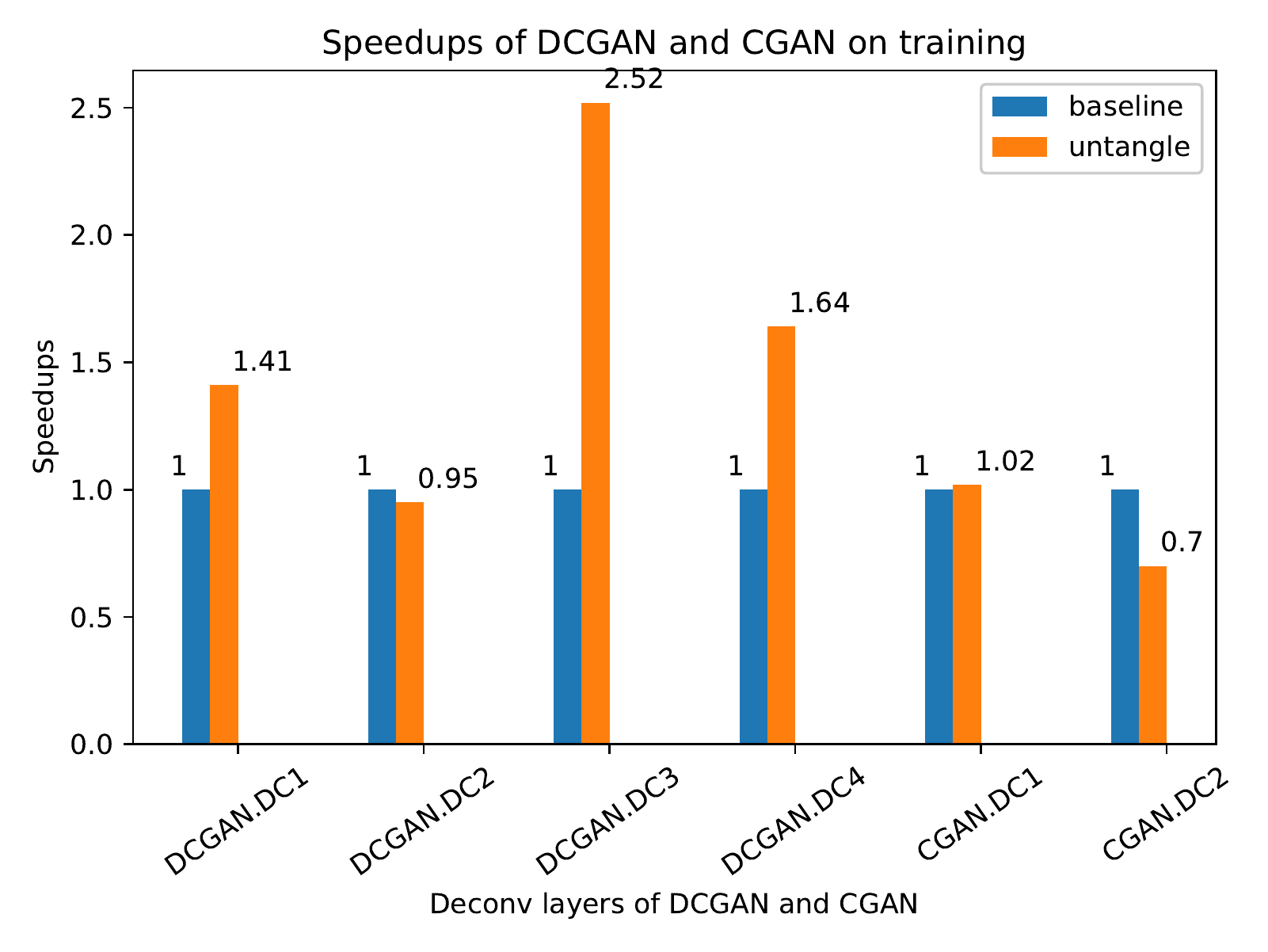}
  \caption{left: the memory access reduction for GANs; right: the speedup of training GANs}
  \label{fig:ma_gan}
\end{figure}

The results of experiments for the memory access reduction by decomposition and untangling are provided in Figure \ref{fig:ma_gan}.

One more thing we want to mention is that untangling technique we applied favors the $C \times N \times R \times S$ memory layout for the transposed kernels and $C \times H \times W$ for the input tensor. This is because elements along $C$ and $N$ dimensions are stored consecutively in these layouts, and this helps with the data fetching in coalescing memory access pattern.

As shown in Figure \ref{fig:ma_gan}, it is obvious that the deeper deconvolution layers are data-bounded, the reduction can be obtained more on the deeper layers since the output tensor becomes larger by the unsampling effects. We achieve a memory access reduction around 30\% to 70\% by only applying untangling technique.

\subsubsection{Speedup in GAN training}
The right side of Figure \ref{fig:ma_gan} plots the speedup of training of GANs. We select several typical layers for the experiments, we want to cover both the cases for dilated derivative maps convolving input tensor and derivative maps stridedly convolving input tensors.

\section{Conclusion}\label{sec:conclusion}
In this paper we presented a set of efficient algorithms and optimizations for deconvolutions, these algorithms are the core components in our deep generative model engine "\textit{HUGE}". We devised them as pervasive as possible to fit on most hardware platforms. \textit{HUGE} really accomplishes the outstanding results for our applications. It shows great improvements in two crucial aspects, computation loads and memory access, respectively. 

%


\medskip

\small

\bibliography{references}



\end{document}